\documentclass[11pt, a4paper, logo, copyright, nonumbering]{map}
\usepackage{CJKutf8}
\usepackage{xargs}  
\usepackage{todonotes}
\usepackage{amsmath}
\usepackage{dsfont}

\usepackage{longtable}

\usepackage{svg}
\usepackage{fontawesome}
\usepackage{array}
\usepackage{tabularx}
\usepackage{latexsym}
\usepackage{graphicx}
\usepackage[utf8]{inputenc} 
\usepackage[utf8]{inputenc} 
\usepackage{amssymb} 
\usepackage{url}            
\usepackage{amsfonts}       
\usepackage{nicefrac}       
\usepackage{microtype}      
\usepackage{xspace}

\usepackage{listings}
\usepackage{makecell}
\usepackage{inconsolata}
\usepackage{multicol}
\usepackage{amstext}

\definecolor{darkblue}{RGB}{84, 112, 198}

\definecolor{lightblue}{rgb}{0.85, 0.95, 1.0}    
\definecolor{lightgreen}{rgb}{0.90, 1.0, 0.90}    
\definecolor{lightorange}{rgb}{1.0, 0.95, 0.85}   
\definecolor{lightpurple}{rgb}{0.95, 0.90, 1.0}   
\definecolor{lightgray}{rgb}{0.97, 0.97, 0.97}    

\usepackage{tikz}
\usetikzlibrary{calc}
\definecolor{battery-empty}{rgb}{0.9, 0.9, 0.9}
\newcommand{\difficultybar}[1]{%
  \begin{tikzpicture}[baseline, scale=0.5, every node/.style={scale=0.8}]
    \foreach \i in {1,2,3,4,5} {
      \ifnum\i>#1
        \draw[fill=battery-empty] (\i*0.5-0.5, 0) rectangle (\i*0.5, 0.25);
      \else
        \pgfmathsetmacro{\colorlevel}{80 - 12*(\i)} 
        \edef\x{\noexpand\draw[fill=blue!\colorlevel!white, opacity=0.9] (\i*0.5-0.5, 0) rectangle (\i*0.5, 0.25);}
        \x
        \draw[blue!50!black] (\i*0.5-0.5, 0) rectangle (\i*0.5, 0.25);
      \fi
    }
    \fill[battery-empty!70] (2.5, 0.08) rectangle (2.6, 0.17);
    \draw[battery-empty!70!black] (2.5, 0.08) rectangle (2.6, 0.17);
  \end{tikzpicture}%
}
\renewcommand{\arraystretch}{0.96}

\definecolor{hidden-draw}{RGB}{20,68,106}
\definecolor{hidden-pink}{RGB}{255,245,247}
\usepackage[edges]{forest}

\usepackage{bm}

\usepackage{natbib}
\usepackage{CJKutf8}
\usepackage{xargs}
\usepackage{todonotes}
\usepackage{multirow}
\usepackage{cleveref}
\usepackage{amsmath}
\usepackage{dsfont}
\usepackage{subcaption}
\usepackage{url}
\usepackage{svg}
\usepackage{fontawesome}
\usepackage{xcolor}
\usepackage{float}

\usepackage[utf8]{inputenc}
\usepackage{booktabs}
\usepackage{graphicx}
\usepackage{array}
\usepackage{tabularx}
\usepackage{xcolor,colortbl}  
\definecolor{boxcolor}{HTML}{d92523} 
\definecolor{bulbcolor}{HTML}{e3b87f} 
\usepackage{url}
\usepackage{ragged2e}   
\usepackage{makecell} 

\usepackage{hyperref}       
\usepackage{url}            
\usepackage{booktabs}       
\usepackage{amsfonts}       
\usepackage{nicefrac}       
\usepackage{microtype}      
\usepackage{xcolor}         
\usepackage{graphicx}
\usepackage{longtable}
\usepackage{array}
\usepackage{pbox}
\usepackage{amsmath}
\usepackage{amssymb}
\usepackage{tabularx}
\usepackage{multirow}
\usepackage{wrapfig}
\usepackage{pifont}
\usepackage{algorithm}
\usepackage{algorithmic}
\usepackage{lipsum}
\usepackage{subcaption}
\usepackage{enumitem}
\usepackage{tikz}
\usepackage{xstring}

\usepackage{color-edits}

\usepackage{booktabs}
\usepackage{multirow}
\usepackage[table]{xcolor}
\usepackage{tabularx}

\usepackage{parskip} 

\usepackage{threeparttable} 

\newcommand{\modelname}{IQuest-Coder-V1} 
\definecolor{rliableolive}{HTML}{BBCC33}
\definecolor{rliableblue}{HTML}{77AADD}
\definecolor{rliablered}{HTML}{f63c44}

\definecolor{rliableolive}{HTML}{BBCC33}
\definecolor{rliableblue}{HTML}{77AADD}
\definecolor{rliablered}{HTML}{f63c44}

\usepackage[most,skins,theorems]{tcolorbox}
\tcbuselibrary{skins,breakable,fitting,hooks}  
\tcbset{
  aibox/.style={
    width=\linewidth,
    top=7pt,
    bottom=2pt,
    colback=rliablered!18!white,
    colframe=black,
    colbacktitle=black,
    enhanced,
    center,
    attach boxed title to top left={yshift=-0.1in,xshift=0.15in},
    boxed title style={boxrule=0pt,colframe=white,},
  }
}
\tcbset{
  aibox2/.style={
    width=\linewidth,
    top=7pt,
    bottom=2pt,
    colback=green!18!white,
    colframe=black,
    colbacktitle=black,
    enhanced,
    center,
    attach boxed title to top left={yshift=-0.1in,xshift=0.15in},
    boxed title style={boxrule=0pt,colframe=white,},
  }
}
\newtcolorbox{AIbox}[2][]{aibox,title=#2,#1}
\newtcolorbox{AIbox2}[2][]{aibox2,title=#2,#1}

\hypersetup{
    colorlinks=true,            
    linkcolor=blue,             
    filecolor=magenta,          
    urlcolor=cyan,              
    citecolor=purple,             
    pdftitle={Overleaf Example},
    pdfpagemode=FullScreen,
}

\definecolor{iquestblue}{HTML}{173C7F}
\definecolor{iquestazure}{HTML}{528FCC}

\renewcommand{\today}{}
\newcommandx{\info}[2][1=]{\todo[linecolor=red,backgroundcolor=red!25,bordercolor=red,#1]{#2}}

\title{
\vspace{-0.2in}
\centering \fontsize{15pt}{16pt}\selectfont
\raisebox{-0.02\height}{ 
        \includegraphics[width=14.5mm]{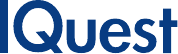}
}-Coder-V1 Technical Report
\vspace{-0.2in}
}

\author{
IQuest Coder Team
\vspace{-5pt}
}

\begin{abstract}
In this report, we introduce the \modelname{} series-(7B/14B/40B/40B-Loop), a new family of code large language models (LLMs).
Moving beyond static code representations, we propose the \texttt{\textbf{code-flow}} multi-stage training paradigm, which captures the dynamic evolution of software logic through different phases of the pipeline. 
Our models are developed through the evolutionary pipeline, starting with the initial pre-training consisting of code facts, repository, and completion data. Following that, we implement a specialized mid-training stage that integrates reasoning and agentic trajectories in 32k-context and repository-scale in 128k-context to forge deep logical foundations.
The models are then finalized with post-training of specialized coding capabilities, which is bifurcated into two specialized paths: the thinking path (utilizing reasoning-driven RL) and the instruct path (optimized for general assistance). 
\modelname{} achieves state-of-the-art performance among competitive models across critical dimensions of code intelligence: agentic software engineering, competitive programming, and complex tool use. 
To address deployment constraints, the \modelname{}-Loop variant introduces a recurrent mechanism designed to optimize the trade-off between model capacity and deployment footprint, offering an architecturally enhanced path for efficacy-efficiency trade-off.
We believe the release of the \modelname{} series, including the complete white-box chain of checkpoints from pre-training bases to the final thinking and instruction models, will advance research in autonomous code intelligence and real-world agentic systems.
\end{abstract}

\begin{document}

\maketitle

\let\oldthefootnote\thefootnote

\begin{figure*}[h!]
    \centering
    \includegraphics[width=0.9\textwidth]{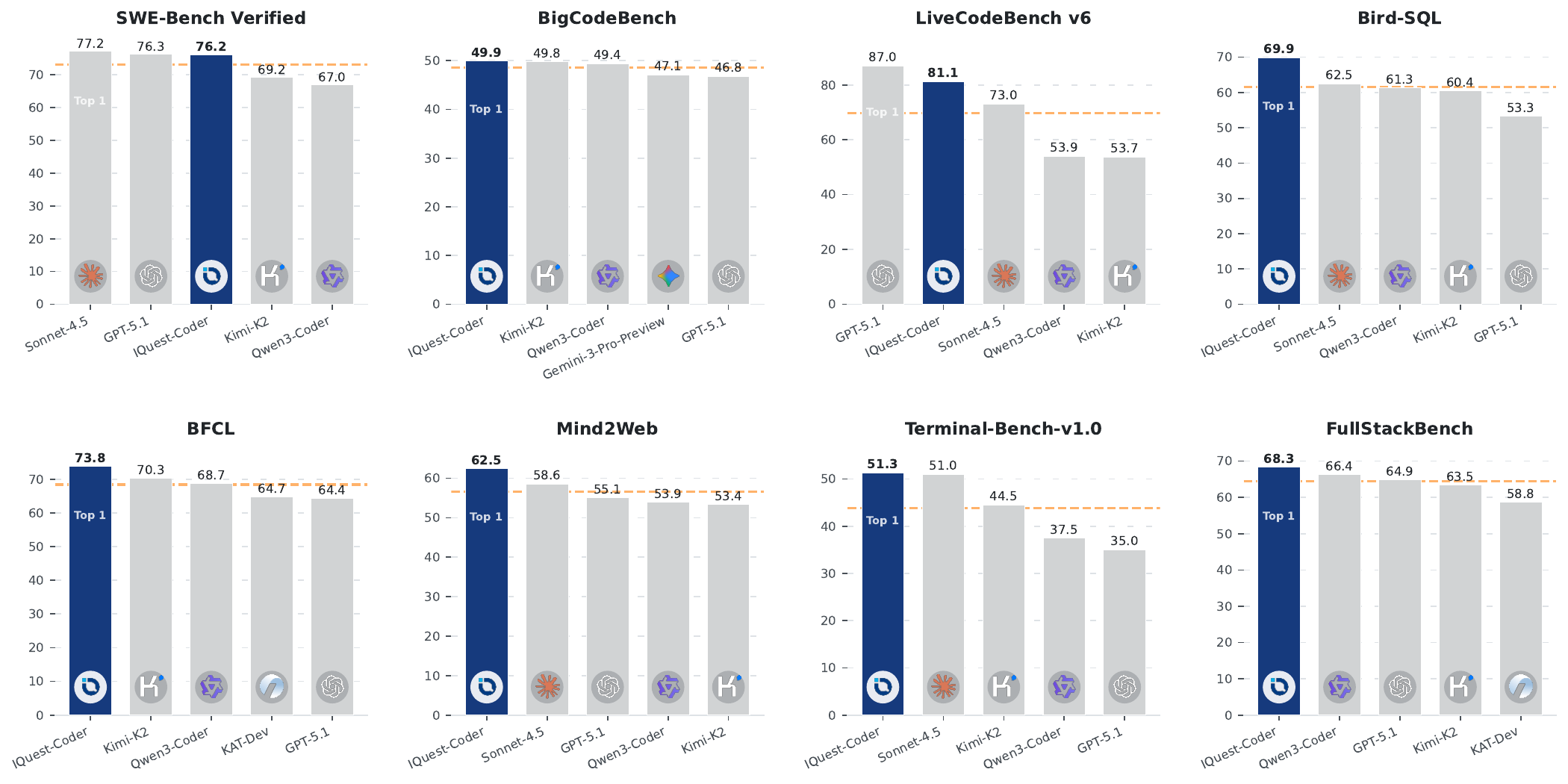}
    \vspace{-5mm}
    \caption{\modelname{} performance across different benchmarks. 
    The score of LiveCodeBench v6 is from \modelname{}-40B-Loop-Thinking model, and the rest are \modelname{}-40B-Loop-Instruct model.
    The orange dash line represents the average score of the selected models.}
    \label{fig:tease_perf}
\end{figure*}

\newpage

\newpage

\section{Introduction}

The current generation of large language models (LLMs) has demonstrated that general-purpose intelligence can be significantly amplified through domain-specific specialization~\cite{yang2025codesurvey}. However, in the field of code intelligence, a wide gap remains between open-weights models and proprietary leaders like Claude 4.5 Sonnet\footnote{https://www.anthropic.com/claude/sonnet}. This gap is most evident in long-horizon reasoning and the ability to navigate complex, multi-file codebases~\cite{swebench}. We introduce \textbf{\modelname{}} series, a family of dense models ranging from 7B to 40B parameters, built to close this gap by maximizing the intelligence density through a structured, multi-phase evolution of logic.

\begin{figure*}[t!]
    \centering
    \includegraphics[width=0.85\textwidth]{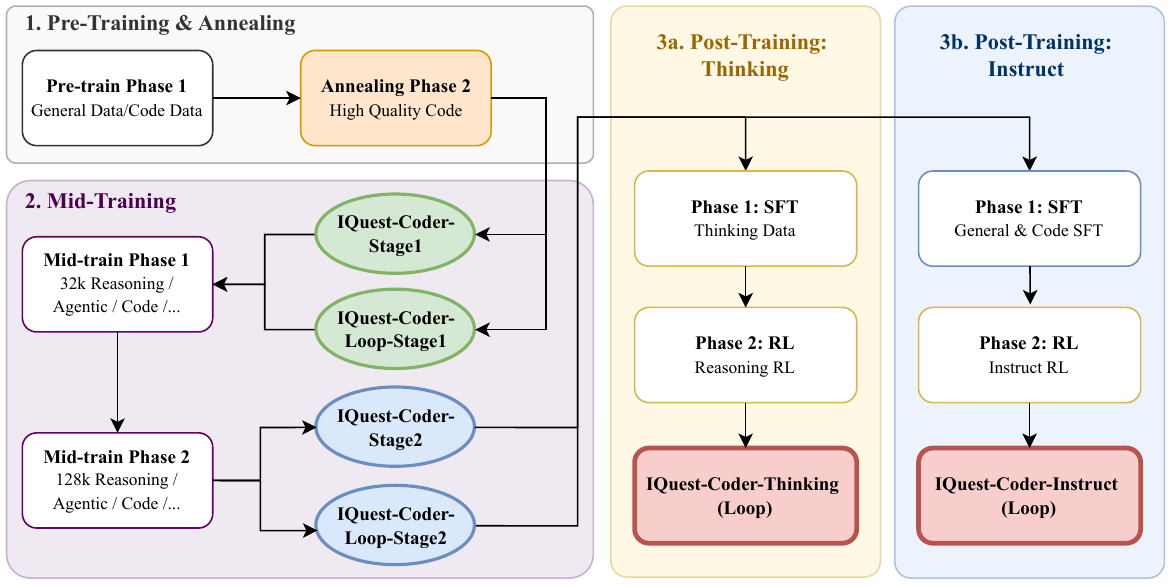}
     \caption{\texttt{\textbf{Code-Flow}} Training pipeline of \modelname{}.}
    \label{fig:pipeline}
\end{figure*}

Our technical contributions are centered around a four-pillar \texttt{\textbf{Code-Flow}} pipeline (\autoref{fig:pipeline}):
\begin{itemize}
    \item Pre-training \& High-Quality Annealing: We begin with a two-stage pre-training process that transitions from stage-1 general data to stage-2 broad code data. This is followed by a targeted annealing phase using high-quality curated code, ensuring the model's base representations are primed for the complex logical tasks that follow.
    \item Dual-Phase Mid-training: To bridge the gap between static knowledge and agentic action, we introduce a dedicated mid-training stage with reasoning, agentic, and long-context coding data.
    \item Bifurcated Post-training: Recognizing that different use cases require different optimization profiles, we offer two distinct post-training paths focusing on instruction tuning and thinking paths.
    \item Efficient Architectures: Our loop model incorporates a recurrent structure to enable iterative computation over complex code segments, providing a scalable architectural path within the constraints of real-world deployment.
\end{itemize}

IQuest-Coder models are developed through a rigorous training methodology that combines large-scale pretraining on extensive code repositories with specialized instruction tuning. Our pretraining corpus encompasses billions of tokens from diverse sources, including public code repositories, technical documentation, and programming-related web content. We employ sophisticated data cleaning and filtering techniques to ensure high-quality training data, implementing both repository-level and file-level processing strategies to capture code structure and context effectively.
The model series demonstrates three key characteristics: (1) Superior Performance: Our flagship IQuest-Coder-40B model achieves state-of-the-art results on major coding benchmarks, demonstrating competitive performance with leading proprietary models. (2) Comprehensive Coverage: With three distinct model sizes ranging from 2B to 40B parameters, IQuest-Coder addresses the diverse needs of the developer community, from resource-constrained edge deployment to high-performance cloud applications. (3) Balanced Capabilities: Beyond code generation, IQuest-Coder maintains strong performance in general language understanding and mathematical reasoning, making it suitable for multi-faceted development tasks.

Through our systematic exploration of the \modelname{} training pipeline, we identified several pivotal findings that offer a deeper understanding of how logical intelligence and agentic capabilities emerge within language models. These insights, derived from extensive ablations of our \texttt{\textbf{code-flow}} data and mid-training strategies, challenge several conventional assumptions in code LLM development:
\begin{itemize}
    \item \textbf{Finding 1}: The repository transition data (the flow of commits) provides a superior signal for task planning compared to training on usual static snapshot files alone.
    \item \textbf{Finding 2}: Injecting 32k reasoning and agentic trajectories after high-quality code annealing—but before post-training—serves as a critical logical scaffold that stabilizes model performance under distribution shifts.
    \item \textbf{Finding 3}: The thinking path (utilizing RL) triggers an emergent ability for autonomous error-recovery in long-horizon tasks (e.g. SWE and code contest tasks) that is largely absent in standard Instruct SFT post-training paths.
\end{itemize}

\begin{figure*}[h!]
    \centering
    \includegraphics[width=1.0\textwidth]{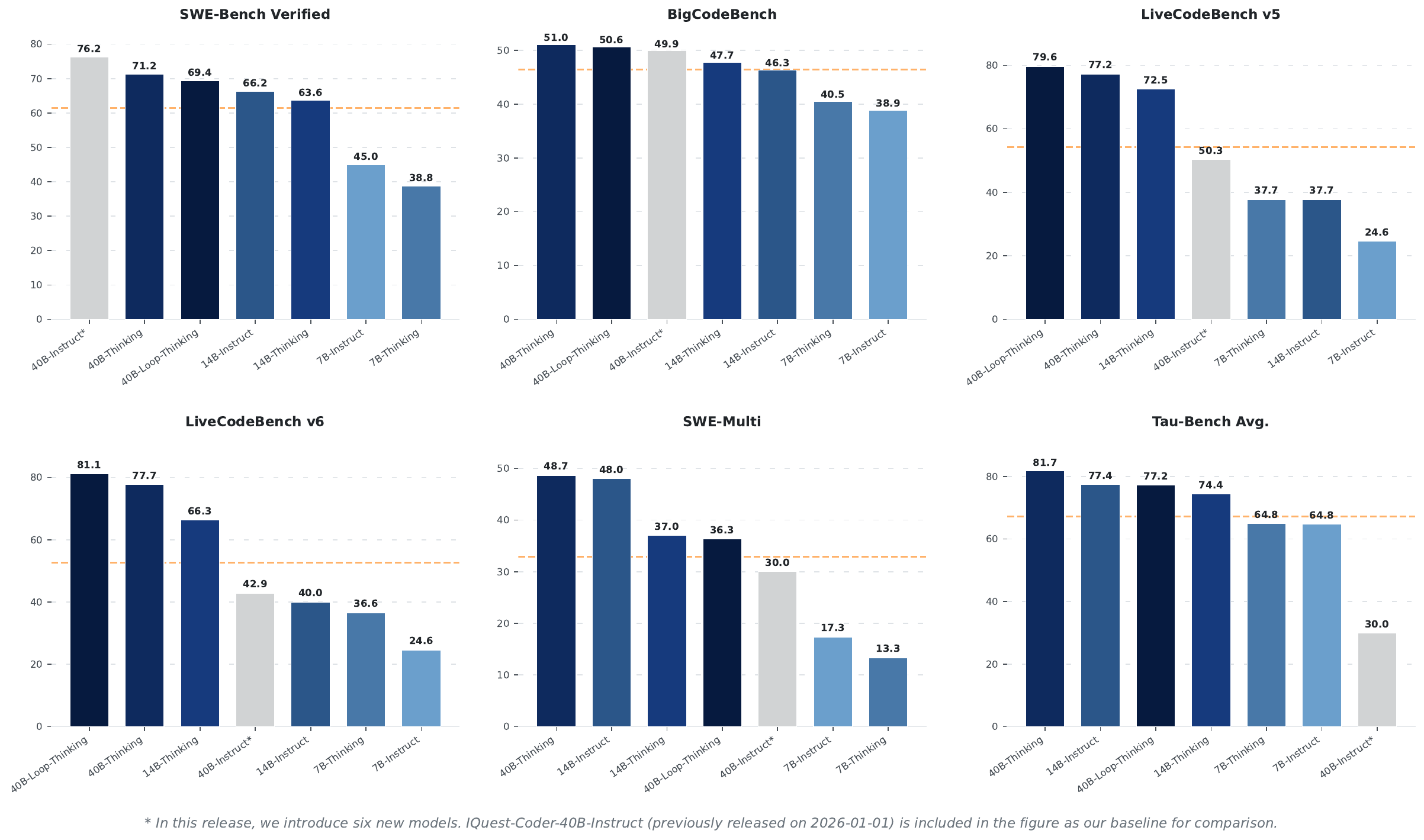}
    \caption{\modelname{} performance across different benchmarks.}
    \label{fig:tease_perf_2}
\end{figure*}

Our post-training process leverages carefully curated datasets covering a wide spectrum of programming paradigms, languages, and real-world coding scenarios. This ensures that IQuest-Coder models can serve as effective coding assistants, capable of understanding complex requirements, generating robust solutions, and providing helpful explanations as revealed in \autoref{fig:tease_perf} and \autoref{fig:tease_perf_2}. We conduct extensive evaluations across popular benchmarks to validate the effectiveness of our approach, with results demonstrating significant improvements over existing open-source alternatives (\emph{ref.} \autoref{sec:evaluation}). 
By releasing the complete evolutionary chain from stage 1 to the final post-training checkpoints, we provide a white-box resource for the community to study the forging of agentic code intelligence.

\section{Model Architecture}

\begin{table*}[t]
  \centering
  \resizebox{1.0\textwidth}{!}{%
    \begin{tabular}{lcccccccc}
      \toprule
      \textbf{Model Size} & \textbf{Layers} &  \textbf{Hidden Size} & \textbf{ Intermediate Size} &
      \textbf{Attention}  &
      \textbf{Max Context} &
      \textbf{Query Heads} &
      \textbf{KV Heads} &
      \textbf{Vocabulary} \\
      \midrule
\multicolumn{9}{c}{\textbf{Base Models (Stage 1)}} \\ \midrule
\modelname{}-7B-Base-Stage1  & 14 & 5120 & 27648 & GQA & 131072 & 40 & 8 & 76800 \\
\modelname{}-14B-Base-Stage1 & 28 & 5120 & 27648 & GQA & 131072 & 40 & 8 & 76800 \\
\modelname{}-40B-Base-Stage1 & 80 & 5120 & 27648 & GQA & 131072 & 40 & 8 & 76800 \\ 
\midrule
\multicolumn{9}{c}{\textbf{Base Models (Stage 2)}} \\ \midrule
\modelname{}-7B-Base  & 14 & 5120 & 27648 & GQA & 131072 & 40 & 8 & 76800\\
\modelname{}-14B-Base & 28 & 5120 & 27648 & GQA & 131072 & 40 & 8 & 76800\\
\modelname{}-40B-Base & 80 & 5120 & 27648 & GQA  & 131072 & 40 & 8 & 76800 \\
\midrule
\multicolumn{9}{c}{\textbf{Instruct Models}} \\ \midrule
\modelname{}-7B-Instruct  & 14 & 5120 & 27648 & GQA & 131072 & 40 & 8 & 76800\\
\modelname{}-14B-Instruct & 28 & 5120 & 27648 & GQA & 131072 & 40 & 8 & 76800\\
\modelname{}-40B-Instruct & 80 & 5120 & 27648 & GQA & 131072 & 40 & 8 & 76800\\
\makecell[l]{\modelname{}-40B-Loop-Instruct \\ (LoopCoder-Instruct)} & 80 & 5120 & 27648 & GQA & 131072 & 40 & 8 & 76800\\ \midrule
\multicolumn{9}{c}{\textbf{Thinking Models}} \\ \midrule
\modelname{}-7B-Thinking  & 14 & 5120 & 27648 & GQA & 131072 & 40 & 8 & 76800\\
\modelname{}-14B-Thinking & 28 & 5120 & 27648 & GQA & 131072 & 40 & 8 & 76800\\
\modelname{}-40B-Thinking & 80 & 5120 & 27648 & GQA & 131072 & 40 & 8 & 76800\\
\bottomrule
    \end{tabular}%
  }%
  \caption{Architecture of \modelname{}.}
  \label{tab:model_architecture}
\end{table*}

\subsection{LoopCoder}

\paragraph{LoopCoder Architecture.}
The LoopCoder architecture employs a loop transformer design where transformer blocks with shared parameters are executed in two fixed iterations. In the first iteration, input embeddings are processed through transformer layers with position-shifted hidden states. During the second iteration, the model computes two types of attention: global attention (where queries from iteration 2 attend to all key-value pairs from iteration 1) and local attention (where queries attend only to preceding tokens within iteration 2 to maintain causality). These two attention outputs are combined using a learned gating mechanism based on query representations, with the gate controlling the weighted mixture of global context refinement and local causal dependencies. This approach differs from the original Parallel Loop Transformer by omitting token-shifting mechanisms and inference-specific optimizations.

\paragraph{LoopCoder Training.}
The training pipeline for LoopCoder consists of three main stages, as illustrated in~\autoref{fig:pipeline}.

\textbf{Stage 1: Pre-Training \& Annealing.}
The training begins with a pre-training phase using a mixture of general data and code data, followed by an annealing phase that focuses on high-quality code corpora. This stage establishes the foundational language understanding and code generation capabilities of the model.

\textbf{Stage 2: Mid-Training.}
The mid-training stage is divided into two phases with progressively increasing context lengths. In Mid-train Phase 1, we train the model on 32k context data comprising reasoning, agentic, and code tasks, yielding \modelname{}-Base-Stage1. In Mid-train Phase 2, we further extend the context length to 128k and continue training on similar data distributions. This phase produces \modelname{}-Base, which serve as the base models for subsequent post-training.

\textbf{Stage 3: Post-Training.}
We develop two variants of LoopCoder through distinct post-training recipes:
\begin{itemize}
    \item \textbf{Thinking Models:} We first perform supervised fine-tuning (SFT) on thinking data that contains explicit reasoning traces, followed by reinforcement learning (RL) optimized for reasoning capabilities. 
    \item \textbf{Instruct Models:} We apply SFT on general and code instruction-following data, then conduct RL training to enhance instruction-following abilities. This produces LoopCoder-Instruct.
\end{itemize}

\subsection{Infra for LoopCoder}
This document describes the three-stage training methodology and infrastructure from LoopCoder. The training progresses from (1) pre-training on general and code data with annealing on high-quality code, to (2) mid-training with progressively longer contexts (32k then 128k) on reasoning, agentic, and code tasks, and finally (3) post-training via two pathways—SFT and RL for either thinking models (with explicit reasoning) or instruct models (for instruction-following). Supporting this multi-million GPU-hour training effort, the infrastructure prioritizes computational efficiency through fused gated attention kernels that reduce memory bandwidth overhead, context parallelism that enables ultra-long context training via point-to-point KV shard transmission with reduced memory costs, and reliability through silent error detection using deterministic re-computation and tensor fingerprint validation to catch hardware failures that don't trigger explicit exceptions.

\section{Pre-training}
We adopt the pre-training guideline~\cite{yang2025scaling} for the code pre-training, which has direct implications for constructing multilingual code corpora. When training tokens are limited, prioritizing mixing syntactically-related PLs can further bring more improvement compared to naively upsampling a single PL. The positive synergy effects suggest that linguistic diversity, particularly when it spans across the code domain, acts as a form of data augmentation that improves model robustness. Taking into account the synergistic effects of different programming languages (PL), we ultimately construct code pre-training data through a reasonable data allocation.

\subsection{Stage1: General Pre-training}

\paragraph{General Corpus Proccessing}
To construct the foundational corpus for IQuest-Coder, we curated a massive dataset primarily sourced from Common Crawl\footnote{https://commoncrawl.org/}. Our pre-processing pipeline begins with a rigorous cleaning stage utilizing regular expressions to remove low-quality noise and non-informative fragments. We ensure data integrity through a hierarchical deduplication strategy, combining exact match filtering with fuzzy deduplication driven by high-dimensional embedding models. 
To safeguard the validity of our evaluations, a comprehensive decontamination procedure is implemented to eliminate any overlaps with common benchmarks. For programming data retrieved from Common Crawl, we perform deep Abstract Syntax Tree (AST) analysis to verify syntactic structure and structural integrity, a critical step for our code-flow training paradigm.
To scale quality control, we train a suite of domain-specific proxy classifiers specialized for general text, code, and mathematics. These proxies are designed to emulate the quality assessment capabilities of much larger models, which provide annotation samples across dimensions such as information density, educational value, and toxic content. Empirical results on validation sets confirm that these small proxy models outperform traditional FastText-based approaches, providing a far more precise signal for selecting high-utility tokens. To enhance the code-related factuality of LLM, we use CodeSimpleQA-Instruct~\cite{codesimpleqa}, a large-scale instruction corpus with 66 million samples, into the pre-training stage. LLMs are adopted to automatically generate factual question-answer pairs from each cluster through a structured pipeline that incorporates explicit constraints to ensure questions are objective, unambiguous, and time-invariant with single correct answers. This approach produces high-quality, objective technical assessments suitable for knowledge evaluation platforms while ensuring time-invariant accuracy and requiring minimal ongoing maintenance.

To construct a dataset suitable for learning repository evolution patterns, we design a triplet construction strategy based on project lifecycle. For each code repository, the system constructs triplets of the form $(\mathcal{R}_{old}, \mathcal{P}, \mathcal{R}_{new})$, where $\mathcal{R}_{old}$ represents the project's code state at a stable development phase, $\mathcal{P}$ denotes the patch information capturing differences between two code states, and $\mathcal{R}_{new}$ represents the code state after a series of development iterations. The starting point selection follows a \textit{project maturity principle}: commits are selected within the 40\%-80\% percentile range of the project lifecycle. This interval corresponds to the mature development phase of the project, where the codebase is relatively stable, avoiding both the uncertainty of early development and the fragmented changes typical of late-stage maintenance. This approach ensures that training data reflects authentic software development patterns. Based on the selected starting point, the system searches forward for appropriate endpoint commits to form complete triplets. The search strategy considers the quality and representativeness of code changes, ensuring that each triplet captures meaningful development iteration processes. This construction method generates training data that maintains the temporal continuity of code evolution while ensuring data diversity and information density, providing a theoretically sound foundational dataset for LLM to learn complex code transformation patterns.



\paragraph{Code Completion}
Code completion is a fundamental capability of code intelligence. This proficiency is primarily enhanced by training on data constructed in the Fill-In-the-Middle (FIM)~\cite{bavarian2022efficienttraininglanguagemodels} format. In the FIM paradigm, a code document is partitioned into three segments: prefix, middle, and suffix. The training objective is to predict the middle content based on the provided prefix and suffix. File-level FIM focuses on individual documents, where the segments are concatenated for training with Fill-In-the-Middle (FIM) pattern. Furthermore, Repo-level FIM extends this approach by incorporating semantically similar code snippets from the same repository as additional context to assist in predicting the middle segment. We primarily employ two strategies for code completion data construction: heuristic-based and multi-level syntax-based construction~\cite{execrepobench}.

The heuristic-based approach consists of two techniques: random boundary splitting and random line splitting. Random boundary splitting partitions code documents at a character-level granularity, which enhances the model's generalization and improves its performance in generating large code blocks or continuing from specific characters. In contrast, random line splitting selects a specific line within the document as the target for completion, which better aligns with typical user interaction patterns. The syntax-based approach leverages the inherent structural properties of source code. By utilizing abstract syntax tree (AST) representations, we extract code segments from various nodes with different characteristics. This method ensures both the randomness of the training data and the structural integrity of the code. We implement several hierarchical levels, including expression-level, statement-level, and function-level. Based on these nodes, we construct multiple PLs and multi-level completion data for both file-level and repo-level tasks, significantly enhancing the diversity of the training samples.The task structure for file-level completion is \texttt{\small <|fim\_prefix|>\{code\_pre\}<|fim\_suffix|>\{code\_suf\}<|fim\_middle|>\{code\_mid\}<|im\_end|>} and the task structure for repository-level completion is
\texttt{\small<|repo\_name|>\{repo\_name\} \\
<|file\_sep|>\{file\_path1\}
\{file\_content1\}
<|file\_sep|>\{file\_path2\}
\{file\_content2\} \\
<|file\_sep|>\{file\_path3\}
<|fim\_prefix|>\{code\_pre\}<|fim\_suffix|>\{code\_suf\}\\<|fim\_middle|>\{code\_fim\}<|im\_end|>}

\subsection{Stage2: Mid-Training}
This mid-training process uses a two-stage approach (Stage 2.1 at 32K context and Stage 2.2 at 128K context) to efficiently scale model capabilities while managing computational costs. Both stages train on the same core data categories: Reasoning QA (math, coding, logic), Agent trajectories, code commits, and file/repository-level fill-in-the-middle (FIM) data. The Reasoning QA component acts as a "reasoning runtime" that encourages structured problem decomposition and consistency checking rather than simple pattern matching, while Agent trajectory data teaches "closed-loop intelligence" by exposing the model to complete action-observation-revision cycles with dense environmental feedback (commands, logs, errors, test results). This combination provides both symbolic reasoning scaffolding and grounded ``code world'' experience, enabling the model to handle long-horizon tasks, recover from errors, and maintain coherent plans across extended contexts, with Stage 2.2 specifically extending these capabilities to repository-level reasoning by incorporating dedicated 128K sequence length samples.

\section{Post-Training}

Post-training transforms pre-trained models into specialized code intelligence systems through supervised fine-tuning and reinforcement learning. This phase uses instructional data spanning code engineering, mathematics, agentic capabilities, and conversation, employing model-in-the-loop synthesis with execution-based verification.

\subsection{Data Construction}
We employ a model-centric framework where frontier LLMs generate training data under rigorous automated verification, using deterministic execution-based validation for objective domains and ensemble mechanisms combining rule-based checks, reward models, and multi-agent debate for subjective domains. Our methodology spans API orchestration, full-stack engineering, competitive programming, code reasoning, text-to-SQL, code editing, terminal benchmarking, repository-scale engineering, tool use, and GUI agents, synthesizing data through techniques like stochastic perturbations, test-driven synthesis, reverse pipeline generation, and multi-stage filtering with automated environment construction. This is followed by large-scale supervised fine-tuning that processes token counts near pre-training scale to inject dense task-specific knowledge, utilizing optimization infrastructure such as aggressive sequence packing, conservative cosine annealing learning rates, and a three-phase curriculum that sequences data by difficulty to ensure stable convergence and superior performance on complex benchmarks.

\subsection{Large-Scale Supervised Fine-Tuning}
Post-training processes match pre-training scale to inject specialized knowledge through optimized infrastructure, including sequence packing with cross-sample masking, cosine learning rate schedules with extended low-rate phases, and three-phase curriculum learning progressing from basic instruction-following to adversarial examples. Quality control ensures only verified samples enter training through comprehensive sandboxed execution, capturing traces and metrics, symbolic mathematical verification, multi-agent debate for subjective evaluation, and aggressive contamination prevention via n-gram matching and MinHash LSH deduplication, prioritizing quality over quantity for improved generalization on complex benchmarks.

\subsection{Multi-Objective Optimization}
This section includes three main components: (1) Alignment tax mitigation through replay buffers, dynamic mixture adaptation, and compositional design to preserve general capabilities while specializing; (2) Reinforcement learning from verifiable feedback using GRPO algorithm with clip-Higher strategy on competition coding tasks, trained on test case pass rates without KL penalties; and (3) SWE-RL framework built on scalable cloud-based sandbox infrastructure that formulates real-world software engineering as interactive RL environments, where agents use tool-based actions across multiple steps and are trained via GRPO with rewards based on test suite passage plus regularization for efficiency, enabling parallel trajectory execution for stable long-horizon code reasoning and debugging capabilities—together yielding emergent capabilities like self-debugging, cross-language transfer, and improved uncertainty calibration.

\definecolor{tablegray}{gray}{0.92}
\begin{table}[t!]
    \centering
    \resizebox{0.75\textwidth}{!}{\begin{tabular}{l | cc cc cc cc cc}
        \toprule
        \multirow{2}{*}{\textbf{Model}} & \multicolumn{2}{c}{\textbf{Python}} & \multicolumn{2}{c}{\textbf{Java}} & \multicolumn{2}{c}{\textbf{TypeScript}} & \multicolumn{2}{c}{\textbf{C\#}} & \multicolumn{2}{c}{\textbf{Average}} \\
        \cmidrule(lr){2-3} \cmidrule(lr){4-5} \cmidrule(lr){6-7} \cmidrule(lr){8-9} \cmidrule(lr){10-11} 
        & \textbf{EM} & \textbf{ES} & \textbf{EM} & \textbf{ES} & \textbf{EM} & \textbf{ES} & \textbf{EM} & \textbf{ES} & \textbf{EM} & \textbf{ES} \\
        \midrule
        \multicolumn{11}{c}{\textbf{6B+ Models}} \\
        \midrule
        DeepSeek-Coder-6.7B-Base & 41.1 & 79.2 & 39.9 & 80.1 & 46.3 & 82.4 & 55.0 & 86.9 & 45.6 & 82.1 \\
        DS-Coder-V2-Lite-Base & 41.8 & 78.3 & 46.1 & 81.2 & 44.6 & 81.4 & 58.7 & 87.9 & 47.8 & 82.2 \\
        CodeQwen1.5-7B & 40.7 & 77.8 & 47.0 & 81.6 & 45.8 & 82.2 & 59.7 & 87.6 & 48.3 & 82.3 \\
        Qwen2.5-Coder-7B & 42.4 & 78.6 & 48.1 & 82.6 & 46.8 & 83.4 & 59.7 & 87.9 & 49.3 & 83.1 \\
        StarCoder2-7B & 10.9 & 63.1 & 8.3 & 71.0 & 6.7 & 76.8 & 7.3 & 72.1 & 8.3 & 70.8 \\
        \midrule
        \multicolumn{11}{c}{\textbf{14B+ Models}} \\
        \midrule
        Qwen2.5-Coder-14B & 47.7 & 81.7 & 54.7 & 85.7 & 52.9 & 86.0 & 66.4 & 91.1 & 55.4 & 86.1 \\
        StarCoder2-15B & 28.2 & 70.5 & 26.7 & 71.0 & 24.7 & 76.3 & 25.2 & 74.2 & 26.2 & 73.0 \\
        \midrule
        \multicolumn{11}{c}{\textbf{20B+ Models}} \\
        \midrule
        DS-Coder-33B-Base & 44.2 & 80.4 & 46.5 & 82.7 & 49.2 & 84.0 & 55.2 & 87.8 & 48.8 & 83.7 \\
        Qwen2.5-Coder-32B & 49.2 & 82.1 & 56.4 & \textbf{86.6} & 54.9 & 87.0 & \textbf{68.0} & \textbf{91.6} & 57.1 & \textbf{86.8} \\
        CodeStral-22B & \textbf{49.3} & \textbf{82.7} & 44.1 & 71.1 & 51.0 & 85.0 & 53.7 & 83.6 & 49.5 & 80.6 \\
        
        \rowcolor{tablegray} \textbf{\modelname{}-40B} & 49.0 & 81.7 & \textbf{57.9} & 86.2 & \textbf{61.9} & \textbf{88.5} & 63.4 & 85.5 & \textbf{57.8} & 85.7 \\
        \bottomrule
    \end{tabular}}
    \caption{Performance comparison on CrossCodeEval Tasks.}
    \label{tab:IQuest_results}
    \vspace{-2mm}
\end{table}

\definecolor{tablegray}{gray}{0.92}
\begin{table*}[t!]
    \centering
    \resizebox{1.0\textwidth}{!}{
    \begin{tabular}{l|cccc|cc|c}
    \toprule
        \multirow{2}{*}{\textbf{Model}} & \multicolumn{4}{c}{\textbf{EvalPlus}} & \multicolumn{2}{|c|}{\textbf{BigCodeBench}} & \multirow{2}{*}{\textbf{FullStackBench}} \\ 
        ~ & \textbf{HumanEval} & \textbf{HumanEval+} & \textbf{MBPP} & \textbf{MBPP+} & \textbf{Full} & \textbf{Hard} & ~  \\ 
        \midrule
        \multicolumn{8}{c}{\textbf{6B+ Models}} \\
        \midrule
        DeepSeek-Coder-V2-Lite-Instruct & 81.1 & 75.6 & 85.2 & 70.6 & 37.8 & 18.9 & 49.4 \\ 
        Qwen2.5-Coder-7B-Instruct & \textbf{87.2} & \textbf{81.7} & 84.7 & 72.2 & 37.8 & 13.5 & 42.2 \\ 
        Seed-Coder-8B-Instruct & 81.1 & 75.6 & \textbf{86.2} & \textbf{73.3} & \textbf{44.6} & \textbf{23.6} & \textbf{55.8} \\ 
        \rowcolor{tablegray} \textbf{\modelname{}-7B-Instruct} & 79.9 & 73.2 & 73.5 & 63.5 & 38.9 & 23.0 & 39.7\\ 
        \rowcolor{tablegray} \textbf{\modelname{}-7B-Thinking} & 76.8 & 70.7 & 73.5 & 63.5 & 40.5 & 19.6 & 32.3\\ 
        \midrule
        \multicolumn{8}{c}{\textbf{13B+ Models}} \\
        \midrule
        Qwen2.5-Coder-14B-Instruct & 62.8 & 59.8 & 88.6 & \textbf{77.2} & 47.0 & 6.1 & 53.1 \\ 
        Qwen3-Coder-30B-A3B-Instruct & \textbf{93.9} & \textbf{87.2} & \textbf{90.7} & \textbf{77.2} & 46.9 & \textbf{27.7} & \textbf{60.9} \\ 
        \rowcolor{tablegray} \textbf{\modelname{}-14B-Instruct} & 83.5 & 78.7 & 79.6 & 68.5 & 46.3 & 26.4 & 48.6\\ 
        \rowcolor{tablegray} \textbf{\modelname{}-14B-Thinking} & 92.7 & 86.0 & 90.5 & 72.0 & \textbf{47.7} & 23.7 & 46.6\\ 
        \midrule
        \multicolumn{8}{c}{\textbf{20B+ Models}} \\
        \midrule
        Deepseek-V3.2 & 93.9 & 88.4 & 93.4 & 77.2 & 48.1 & 27.0 & 64.9 \\ 
        Qwen2.5-Coder-32B-Instruct & 93.3 & 86.6 & 90.2 & 77.8 & 48.0 & 24.3 & 57.4 \\  
        Qwen3-235B-A22B-Instruct-2507 & 96.3 & 91.5 & 92.3 & 77.8 & 47.4 & 25.7 & 62.7 \\ 
        Qwen3-235B-A22B-Thinking-2507 & \textbf{98.8} & \textbf{93.3} & 95.5 & 81.5 & 44.1 & 23.0 & - \\ 
        Qwen3-Coder-480B-A35B-Instruct & 97.6 & 92.7 & 94.2 & 80.2 & 49.4 & 27.7 & 66.4 \\ 
        Kimi-Dev-72B & 93.3 & 86.0 & 79.6 & 68.8 & 45.4 & 31.8 & 38.6 \\ 
        Kimi-K2-Instruct-0905 & 94.5 & 89.6 & 91.8 & 74.1 & 49.8 & 30.4 & 63.5 \\ 
        Kimi-K2-Thinking & 98.2 & 92.7 & \textbf{97.4} & \textbf{82.3} & 46.8 & 28.4 & - \\ 
        KAT-Dev & 90.9 & 86.6 & 89.4 & 76.2 & 46.2 & 25.7 & 58.8 \\ 
        KAT-Dev-72B-Exp & 88.4 & 81.7 & 85.2 & 69.3 & 48.3 & 26.4 & 52.9 \\ 
        GLM-4.7 & 87.2 & 79.9 & 90.5 & 75.7 & 45.7 & 26.4 & 70.2 \\ 
        \rowcolor{tablegray} \textbf{\modelname{}-40B-Instruct} & 96.3 & 90.2 & 91.8 & 77.8 & \textbf{54.2} & \textbf{33.1} & \textbf{71.4} \\ 
        \rowcolor{tablegray} \textbf{\modelname{}-40B-Thinking} & 93.9 & 87.8 & 91.0 & 75.1 & 51.1 & 29.1 & 54.8 \\ 
        \rowcolor{tablegray} \textbf{\modelname{}-40B-Loop-Instruct} & 97.6 & 91.5 & 92.9 & 77.2 & 49.9 & 27.7 & 68.3 \\ 
        \rowcolor{tablegray} \textbf{\modelname{}-40B-Loop-Thinking} & 97.6 & 89.6 & 91.0 & 76.2 & 50.6 & 29.7 & 59.5\\
        \midrule
        \multicolumn{8}{c}{\textbf{Closed-APIs Models}} \\
        \midrule
        Gemini-3-Flash-preview & 88.4 & 84.8 & 92.3 & 79.1 & 44.5 & 25.6 & - \\ 
        Gemini-3-Pro-preview & 100.0 & 94.5 & 71.2 & 64.8 & 47.1 & 25.0 & - \\ 
        Claude-Opus-4.5 & 98.8 & 93.3 & 96.8 & 83.9 & 53.3 & 35.1 & 72.3 \\ 
        Claude-Sonnet-4.5 & 98.8 & 93.3 & 95.2 & 82.3 & 51.4 & 29.1 & 69.7 \\ 
        GPT-5.1 & 97.0 & 90.0 & 92.6 & 72.2 & 46.8 & 29.1 & 64.9 \\ 
    \bottomrule
    \end{tabular}
    }
    \label{tab:code_generation_1}
    \caption{Performance comparison on code generation tasks.}
\end{table*}

\definecolor{tablegray}{gray}{0.92}
\begin{table*}[!h]
    \centering
    \resizebox{0.6\textwidth}{!}{
    \begin{tabular}{l|cc|cc}
    \toprule
        \multirow{2}{*}{\textbf{Model}} & \multicolumn{2}{c}{\textbf{CruxEval}} & \multicolumn{2}{c}{\textbf{LiveCodeBench}}  \\ 
        ~ & \textbf{Input-COT} & \textbf{Output-COT} & \textbf{V5} & \textbf{V6} \\ 
        \midrule
        \multicolumn{5}{c}{\textbf{6B+ Models}} \\
        \midrule
        DeepSeek-Coder-V2-Lite-Instruct & 57.1 & 56.2 & 13.2 & 19.4 \\ 
        Qwen2.5-Coder-7B-Instruct & \textbf{66.9} & 66.0 & 14.4 & 18.9 \\ 
        Seed-Coder-8B-Instruct & 62.0 & 66.6 & 19.2 & 22.3 \\ 
        \rowcolor{tablegray} \textbf{\modelname{}-7B-Instruct} & 45.8 & 54.2 & 24.6 & 24.6 \\ 
        \rowcolor{tablegray} \textbf{\modelname{}-7B-Thinking} & 57.6 & \textbf{81.5} & \textbf{37.7} & \textbf{36.6} \\ 
        \midrule
        \multicolumn{5}{c}{\textbf{13B+ Models}} \\
        \midrule
        Qwen2.5-Coder-14B-Instruct & 75.6 & 79.2 & 22.8 & 24.6 \\ 
        Qwen3-Coder-30B-A3B-Instruct & 76.9 & 80.5 & 43.1 & 36.0 \\
        \rowcolor{tablegray} \textbf{\modelname{}-14B-Instruct} & 52.6 & 57.6 & 37.7 & 40.0 \\ 
        \rowcolor{tablegray} \textbf{\modelname{}-14B-Thinking} & \textbf{80.5} & \textbf{90.6} & \textbf{72.5} & \textbf{66.3}\\ 
        \midrule
        \multicolumn{5}{c}{\textbf{20B+ Models}} \\
        \midrule
        DeepSeek-v3.2 & 82.1 & \textbf{94.2} & - & 83.3 \\ 
        Qwen2.5-Coder-32B-Instruct & 78.8 & 84.0 & 30.5 & 27.4 \\  
        Qwen3-235B-A22B-Instruct-2507 & 62.0 & 89.5 & 53.9 & 51.8 \\ 
        Qwen3-235B-A22B-Thinking-2507 & 15.2 & 46.9 & \textbf{80.2} & 74.1 \\ 
        Qwen3-Coder-480B-A35B-Instruct & 87.1 & 90.4 & 48.6 & 53.9 \\  
        Kimi-Dev-72B & 33.0 & 64.2 & 46.1 & 40.0 \\
        Kimi-K2-Instruct-0905 & 86.8 & 89.5 & 52.1 & 53.7 \\ 
        Kimi-K2-Thinking & 92.2 & 86.2 & - & 83.1 \\ 
        KAT-Dev & 42.5 & 65.1 & 32.9 & 32.6 \\ 
        KAT-Dev-72B-Exp & 71.4 & 81.1 & 13.8 & 16.0 \\ 
        GLM-4.7 & 65.6 & 81.2 & - & \textbf{84.9} \\ 
        \rowcolor{tablegray} \textbf{\modelname{}-40B-Instruct} & \textbf{93.5} & 87.0 & 55.7 & 46.9 \\ 
        \rowcolor{tablegray} \textbf{\modelname{}-40B-Thinking} & 87.4 & 94.0 & 77.3 & 77.7 \\ 
        \rowcolor{tablegray} \textbf{\modelname{}-40B-Loop-Instruct} & 91.1 & 85.5 & 48.6 & 48.5 \\ 
        \rowcolor{tablegray} \textbf{\modelname{}-40B-Loop-Thinking} & 76.5 & 75.2 & 79.6 & 81.1 \\ 
        \midrule
        \multicolumn{5}{c}{\textbf{Closed-APIs Models}} \\
        \midrule
        Gemini-3-Flash-preview & 96.5 & 97.6 & - & 90.8 \\ 
        Gemini-3-Pro-preview & 98.8 & 99.1 & - & 91.7 \\ 
        Claude-Opus-4.5 & 98.4 & 98.0 & - & 87.1 \\ 
        Claude-Sonnet-4.5 & 96.2 & 96.2 & - & 73.0 \\
        GPT-5.1 & 70.8 & 71.1 & - & 87.0 \\ 
\bottomrule
    \end{tabular}
    }
    \label{tab:code_generation_3}
    \caption{Performance comparison on Code Reasoning Evaluation.}
\end{table*}

\section{Evaluation}\label{sec:evaluation}
\subsection{Baselines}
In our evaluation, we compare our model against a broad set of state-of-the-art code-focused language models covering instruction-tuned, base, and reasoning-enhanced variants. The baselines span leading closed-source and open-source systems known for strong performance on programming and reasoning tasks, including representative models from Anthropic (Claude 4.5), OpenAI (GPT-5.1), Google (Gemini 3), Alibaba (Qwen and Qwen-Coder series), DeepSeek (Coder and V3 series), Mistral (CodeStral), Moonshot (Kimi), ZhiPu (GLM), Kuaishou (Kwaipilot/KAT), and BigCode (StarCoder2). These models cover a wide parameter range and different tuning strategies, ensuring that our comparison reflects current capability boundaries in code generation, understanding, and complex task execution.

\subsection{Experiments on Base Models}

\subsubsection{Code Completion}
We evaluate cross-file code completion on CrossCodeEval~\cite{ding2023crosscodeevaldiversemultilingualbenchmark}, a multilingual benchmark encompassing Python, Java, TypeScript, and C\#. This benchmark explicitly targets repository-level completion scenarios, serving as a core metric for assessing the fundamental capabilities of code LLMs in leveraging cross-file context.

\subsection{Evaluation on Instruct Models and Reasoning model}

\subsubsection{Code Generation}

Across a wide range of code-generation evaluations, our model achieves consistently strong performance. We validate functional correctness and robustness using EvalPlus~\cite{evalplus} (including HumanEval+ and MBPP+ with substantially expanded test suites), and measure compositional, library-intensive problem solving on BigCodeBench~\cite{zhuo2024bigcodebench}. We further demonstrate broad full-stack capability on FullStackBench~\cite{liu2024fullstackbench}, and strong results under contamination-aware, continuously refreshed testing on LiveCodeBench~\cite{jain2024livecodebench}.

\subsubsection{Code Reasoning}
We further evaluate code reasoning with CRUXEval~\cite{gu2024cruxeval}, which tests both forward execution (Input-to-Output, I2O) and inverse inference (Output-to-Input, O2I) over 800 concise Python functions. Our model performs strongly on I2O and also shows clear gains on the more challenging O2I setting, indicating improved ability to reason about code behavior beyond surface-level execution and to solve inverse constraints implied by a target return value.

\definecolor{tablegray}{gray}{0.92}
\begin{table*}[h]
\centering
\resizebox{0.45\textwidth}{!}{
\begin{tabular}{l|cc}
\toprule
\multirow{2}{*}{\textbf{Model}} & \multicolumn{2}{c}{\textbf{Mercury}} \\
~ & \textbf{Beyond@1} & \textbf{Pass@1} \\
\midrule

\multicolumn{3}{c}{\textbf{6B+ Models}} \\
\midrule
DeepSeek-Coder-V2-Lite-Instruct & 76.8 & 91.4 \\
Qwen2.5-Coder-7B-Instruct & 69.9 & 84.8 \\
Seed-Coder-8B-Instruct & \textbf{78.5} & \textbf{93.8} \\
\rowcolor{tablegray}\textbf{\modelname{}-7B-Instruct} & 42.1 & 50.4 \\
\rowcolor{tablegray}\textbf{\modelname{}-7B-Thinking} & 43.2 & 53.5 \\

\midrule
\multicolumn{3}{c}{\textbf{13B+ Models}} \\
\midrule
Qwen2.5-Coder-14B-Instruct & 76.7 & 88.3 \\
Qwen3-Coder-30B-A3B-Instruct & \textbf{81.1} & \textbf{95.3} \\
\rowcolor{tablegray}\textbf{\modelname{}-14B-Instruct} & 63.3 & 76.2 \\
\rowcolor{tablegray}\textbf{\modelname{}-14B-Thinking} & 62.0 & 74.2 \\

\midrule
\multicolumn{3}{c}{\textbf{20B+ Models}} \\
\midrule
DeepSeek-v3.2 & 81.6 & \textbf{96.9} \\
Qwen2.5-Coder-32B-Instruct & 79.1 & 96.1 \\
Qwen3-235B-A22B-Instruct-2507 & 80.4 & \textbf{96.9} \\
Qwen3-235B-A22B-Thinking-2507 & 61.2 & 70.3 \\
Qwen3-Coder-480B-A35B-Instruct & 80.2 & 96.1 \\
Kimi-Dev-72B & 59.1 & 69.5 \\
Kimi-K2-Instruct-0905 & 76.1 & 90.6 \\
Kimi-K2-Thinking & 73.0 & 85.2 \\
KAT-Dev & 75.1 & 89.1 \\
KAT-Dev-72B-Exp & 79.0 & 94.5 \\
GLM-4.7 & 74.1 & 86.7 \\
\rowcolor{tablegray}\textbf{\modelname{}-40B-Instruct} & \textbf{83.6} & 95.3 \\
\rowcolor{tablegray}\textbf{\modelname{}-40B-Thinking} & 71.1 & 83.2 \\
\rowcolor{tablegray}\textbf{\modelname{}-40B-Loop-Instruct} & 82.2 & 94.1 \\
\rowcolor{tablegray}\textbf{\modelname{}-40B-Loop-Thinking} & 79.6 & 94.9 \\
\midrule
\multicolumn{3}{c}{\textbf{Closed-APIs Models}} \\
\midrule
Gemini-3-Flash-preview & 78.4 & 89.5 \\
Gemini-3-Pro-preview & 83.1 & 96.1 \\
Claude-Opus-4.5 & 82.9 & 96.9 \\
Claude-Sonnet-4.5 & 82.5 & 97.7 \\
GPT-5.1 & 81.9 & 96.1 \\

\bottomrule
\end{tabular}
}
\caption{Performance comparison on code efficiency task.}
\label{tab:code_generation_9}
\vspace{-2mm}
\end{table*}

\subsubsection{Code Efficiency}
We assess code efficiency with Mercury~\cite{du2024mercurycodeefficiencybenchmark}, which evaluates Code LLMs beyond functional correctness by measuring runtime on natural-language-to-code tasks. Mercury contains 256 Python problems across multiple difficulty levels, each with a test-case generator and a set of real-world reference solutions that together define an empirical runtime distribution per task. The benchmark further proposes the percentile-based \emph{Beyond} metric, which reweights Pass by relative runtime to jointly capture correctness and efficiency. Our model achieves strong Mercury results, indicating that it can produce solutions that are not only correct but also competitive in runtime under this distribution-based evaluation.

\subsubsection{Text to SQL}
Our model also performs strongly on cross-domain Text-to-SQL benchmarks that stress generalization to unseen schemas and realistic database settings. On Spider~\cite{2018spider}, which uses a database-level train--test split to evaluate schema linking and structurally correct SQL generation with complex constructs, and on BIRD~\cite{2023bird}, which further emphasizes value grounding from database contents, real-world database scale, and execution-related practicality, our model achieves competitive results, indicating robust semantic parsing and reliable query generation in both schema-centric and content-grounded scenarios.

\definecolor{tablegray}{gray}{0.92}
\begin{table*}[t]
    \centering
    \resizebox{0.7\textwidth}{!}{
    \begin{tabular}{l|cc}
    \toprule 
        \multirow{2}{*}{\textbf{Model}} & \textbf{Bird} & \textbf{Spider} \\ 
        ~ & \textbf{Execution Accuracy} & \textbf{Execution Accuracy} \\ 
    \midrule

    \multicolumn{3}{c}{\textbf{6B+ Models}} \\
    \midrule
        DeepSeek-Coder-V2-Lite-Instruct & 41.6 & 72.4 \\ 
        Qwen2.5-Coder-7B-Instruct & \textbf{53.1} & \textbf{79.8} \\ 
        Seed-Coder-8B-Instruct & 44.7 & 72.7 \\ 
        \rowcolor{tablegray}\textbf{\modelname{}-7B-Instruct} & 37.7 & 67.7 \\
        \rowcolor{tablegray}\textbf{\modelname{}-7B-Thinking} & 30.5 & 56.4 \\

    \midrule
    \multicolumn{3}{c}{\textbf{13B+ Models}} \\
    \midrule
        Qwen2.5-Coder-14B-Instruct & \textbf{59.1} & \textbf{81.3} \\ 
        Qwen3-Coder-30B-A3B-Instruct & 59.0 & 80.9 \\ 
        \rowcolor{tablegray}\textbf{\modelname{}-14B-Instruct} & 50.0 & 75.8 \\
        \rowcolor{tablegray}\textbf{\modelname{}-14B-Thinking} & 46.9 & 69.3 \\

    \midrule
    \multicolumn{3}{c}{\textbf{20B+ Models}} \\
    \midrule
        DeepSeek-v3.2 & 52.6 & 77.9 \\ 
        Qwen2.5-Coder-32B-Instruct & 62.1 & 83.9 \\ 
        Qwen3-235B-A22B-Instruct-2507 & 62.8 & 81.1 \\ 
        Qwen3-235B-A22B-Thinking-2507 & 35.2 & 42.6 \\ 
        Qwen3-Coder-480B-A35B-Instruct & 61.3 & 81.2 \\ 
        Kimi-K2-Instruct-0905 & 60.4 & 81.1 \\ 
        Kimi-K2-Thinking & 40.6 & 49.6 \\ 
        KAT-Dev & 52.2 & 77.6 \\ 
        KAT-Dev-72B-Exp & 35.2 & 60.3 \\ 
        GLM-4.7 & 46.5 & 62.4 \\ 
        \rowcolor{tablegray}\textbf{\modelname{}-40B-Instruct} & \textbf{70.5} & \textbf{92.2} \\
        \rowcolor{tablegray}\textbf{\modelname{}-40B-Thinking} & 53.6 & 78.1 \\
        \rowcolor{tablegray}\textbf{\modelname{}-40B-Loop-Instruct} & 69.9 & 84.0 \\
        \rowcolor{tablegray}\textbf{\modelname{}-40B-Loop-Thinking} & 54.8 & 77.8 \\

    \midrule
    \multicolumn{3}{c}{\textbf{Closed-APIs Models}} \\
    \midrule
        Gemini-3-Flash-preview & 66.6 & \textbf{87.2} \\ 
        Gemini-3-Pro-preview & \textbf{67.5} & 87.0 \\ 
        Claude-Opus-4.5 & 66.0 & 76.0 \\ 
        Claude-Sonnet-4.5 & 62.5 & 80.1 \\ 
        GPT-5.1 & 53.3 & 77.6 \\ 

\bottomrule
    \end{tabular}
    }
    \caption{
     Performance comparison on Text2SQL Tasks.
    }
    \label{tab:code_generation_4}
\end{table*}

\subsubsection{Agentic Coding Tasks}
We further evaluate our model in agentic, end-to-end software workflows where success depends on correct tool use, long-horizon planning, and tight interaction with the execution environment. Terminal-Bench~\cite{tbench2025} measures whether an agent can reliably complete realistic terminal workflows (for example, building software from source, configuring services, managing dependencies, and debugging) inside containerized sandboxes with automated verification, while also standardizing execution via its runner for reproducible leaderboard evaluation. In parallel, SWE-bench~\cite{jimenez2024swebench} targets real-world software engineering by requiring models to produce patches from issue descriptions that turn failing repositories into passing ones under unit-test verification; SWE-bench Verified further improves reliability with 500 curated instances evaluated in a standardized Docker environment, where our model achieves a score of 76.2.

\definecolor{tablegray}{gray}{0.92}

\begin{table*}[h]
    \centering
    \small
    \setlength{\tabcolsep}{3.2pt}
    \renewcommand{\arraystretch}{1.05}
    \resizebox{1.0\textwidth}{!}{
    \begin{tabular}{l|ccc|cc}
    \toprule
        \multirow{2}{*}{\textbf{Model}} &
        \multicolumn{3}{c|}{\textbf{Agentic Coding}} &
        \multicolumn{2}{c}{\textbf{General Tool Use}} \\
        & \textbf{Terminal-Bench} & \textbf{Terminal-Bench (2.0)} & \textbf{SWE-Verified}
        & \textbf{Mind2Web} & \textbf{BFCL V3} \\
    \midrule

        \multicolumn{6}{c}{\textbf{6B+ Models}} \\
    \midrule
        DeepSeek-Coder-V2-Lite-Instruct & 5.0 & 0.0 & -   & 26.7 & - \\
        Qwen2.5-Coder-7B-Instruct       & 6.3 & 0.0 & -   & 38.4 & \textbf{54.2} \\
        Seed-Coder-8B-Instruct          & 7.5 & 2.5 & -   & 38.2 & - \\
        \rowcolor{tablegray}\textbf{\modelname{}-7B-Instruct} & \textbf{22.5} & \textbf{11.2} & \textbf{45.0} & \textbf{40.5} & 34.0 \\
        \rowcolor{tablegray}\textbf{\modelname{}-7B-Thinking} & 21.3 & 6.9 & 38.8 & 10.8 & 43.3 \\
    \midrule
        \multicolumn{6}{c}{\textbf{13B+ Models}} \\
    \midrule
        Qwen2.5-Coder-14B-Instruct      & 8.8  & 0.0  & -    & 42.7 & 59.9 \\
        Qwen3-Coder-30B-A3B-Instruct    & 23.8 & \textbf{23.8} & 51.9 & 36.1 & \textbf{63.4} \\
        \rowcolor{tablegray}\textbf{\modelname{}-14B-Instruct} & \textbf{36.3} & 16.9 & \textbf{66.2} &\textbf{47.1} & 55.1 \\
        \rowcolor{tablegray}\textbf{\modelname{}-14B-Thinking} & 26.3 & 14.1 & 63.6 & 28.7 & 53.6 \\
    \midrule
        \multicolumn{6}{c}{\textbf{20B+ Models}} \\
    \midrule
        DeepSeek-v3.2                   & 23.8 & \textbf{46.4} & 73.1 & 47.2 & 68.8 \\
        Qwen2.5-Coder-32B-Instruct      & 5.0  & 4.5  & -    & 32.5 & 62.3 \\
        Qwen3-235B-A22B-Instruct-2507   & 15.0 & 13.5 & 45.2 & 49.0 & 71.2 \\
        Qwen3-235B-A22B-Thinking-2507   & 8.8  & 3.4  & 44.6 & 43.2 & 71.9 \\
        Qwen3-Coder-480B-A35B-Instruct  & 37.5 & 23.6 & 67.0 & 54.0 & 68.7 \\
        Kimi-Dev-72B                    & -    & 2.3  & 60.4 & -    & 55.5 \\
        Kimi-K2-Instruct-0905           & 44.5 & 27.8 & 69.2 & 53.4 & 70.3 \\
        Kimi-K2-Thinking                & 47.1 & 33.7 & 71.3 & 55.7 & - \\
        KAT-Dev                         & 17.5 & 10.1 & 62.4 & 33.7 & 64.7 \\
        KAT-Dev-72B-Exp                 & 21.3 & 7.9  & 74.6 & -    & - \\
        GLM-4.7                         & 36.3 & 41.0 & 73.8 & 53.7 & 64.8 \\
        \rowcolor{tablegray}\textbf{\modelname{}-40B-Instruct}      & \textbf{52.5} & 33.0 & 70.4 & \textbf{64.3} & 51.7 \\
        \rowcolor{tablegray}\textbf{\modelname{}-40B-Thinking} & 30.0 & 22.3 & 71.2 & 47.6 & 64.2 \\
        \rowcolor{tablegray}\textbf{\modelname{}-40B-Loop-Instruct} & 51.3 & 33.0 & \textbf{76.2} & 62.5 & \textbf{73.9} \\
        \rowcolor{tablegray}\textbf{\modelname{}-40B-Loop-Thinking} & 30.0 & 18.8 & \textbf{76.2} & 62.5 & \textbf{73.9} \\

    \midrule
        \multicolumn{6}{c}{\textbf{Closed-APIs Models}} \\
    \midrule
        Gemini-3-Flash-preview          & 53.8 & 47.6 & 78.0 & 60.6 & - \\
        Gemini-3-Pro-preview            & 46.3 & 54.2 & 76.2 & 60.3 & 78.2 \\
        Claude-Opus-4.5                 & 47.5 & 59.3 & 80.9 & 57.9 & 78.9 \\
        Claude-Sonnet-4.5               & 51.0 & 50.0 & 77.2 & 58.6 & 77.7 \\
        GPT-5.1                         & 35.0 & 47.6 & 76.3 & 55.1 & 64.4 \\
    \bottomrule
    \end{tabular}
    }
    \caption{Combined performance on agentic coding tasks (Terminal-Bench, Terminal-Bench 2.0, SWE-Verified) and general tool-use tasks (Mind2Web, BFCL V3).}
    \label{tab:agentic_combined}
\end{table*}

\subsubsection{Other Agentic Tasks}
Beyond coding-centric agents, we additionally evaluate general tool-use and interactive decision making across web, API, and conversational-agent settings. Mind2Web~\cite{deng2023mind2web} targets generalist web agents that must follow natural-language instructions to complete open-ended tasks on real websites, stressing cross-site generalization and long-horizon UI interaction. BFCL~\cite{patil2025bfcl} tests tool-use across heterogeneous programming and API settings (for example, Java, JavaScript, Python, SQL, and REST APIs), with successive versions increasing realism from broad coverage (v1) to real tool execution (v2), multi-turn multi-step function calling (v3), and holistic agent evaluation that emphasizes autonomous planning and sequential decision making (v4). Finally, $\tau$-bench~\cite{yao2024tau} evaluates conversational agents that must interact naturally with users while following policy constraints, and its extension $\tau^2$-bench further introduces dual-control environments where both the agent and the user can act on a shared world via tools, enabling fine-grained diagnosis of failures in reasoning versus coordination.

\subsubsection{Safety Evaluation}
We adopt the Tulu 3 benchmarking suite~\cite{tulu3} to evaluate safety boundaries, balancing two objectives: maximizing refusals on harmful prompts while minimizing over-refusal on benign inputs in XSTest~\cite{rottger2024xstest} and WildGuardTest~\cite{han2024wildguard}. Response validity is adjudicated by the WildGuard model~\cite{han2024wildguard}, and we report macro-averaged accuracy across all benchmarks, where higher scores indicate better overall safety behavior. Concretely, we evaluate refusals on BeaverTails~\cite{ji2023beavertails} (1,483 harmful prompts), HarmBench~\cite{mazeika2024harmbench} (300 examples from Standard, Contextual, and Copyright subsets), Do-Anything-Now~\cite{shen2024anything} (300 DAN-templated malicious prompts), Do-not-Answer~\cite{wang2023not} (939 harmful prompts), TrustLLM~\cite{huang2024trustllm} (1,400 jailbreak prompts), and WildGuardTest~\cite{han2024wildguard} (780 harmful prompts within 1,725 items), reporting Refusal Rate (RTA) based on whether WildGuard classifies the response as a refusal; for XSTest~\cite{rottger2024xstest}, we report aggregate accuracy by requiring refusals on unsafe prompts and compliance on adversarial benign prompts.

 \begin{table*}[t!]
     \centering
     \resizebox{1.0\textwidth}{!}{
         \begin{tabular}{l|cccccccc}
         \toprule
         \textbf{Model} & \textbf{BeaverTails} & \textbf{HarmBench} & \textbf{Do-Anything-Now} & \textbf{Do-not-Answer} & \textbf{TrustLLM} & \textbf{WildGuardTest} & \textbf{XSTest} & \textbf{Overall} \\
         \midrule
        
         Qwen2.5-Coder-32B-Instruct & 68.0 & 47.5 & 69.7 & 53.7 & 70.2 & 73.9 & 90.6 & 67.7  \\
         Qwen3-Coder-480B-A35B-Instruct      & 70.5 & 94.2 & 95.7 & \textbf{69.9} & \textbf{88.4} & 85.0 & 90.1 & 84.8  \\
        
         \rowcolor{tablegray}\textbf{\modelname{}-40B-Instruct}          & 67.5 & 57.3 & 63.3 & 53.9 & 65.0 & 78.1 & 89.3 & 67.8  \\
         \rowcolor{tablegray}\textbf{\modelname{}-40B-Thinking} & \textbf{76.7} & \textbf{94.8} & \textbf{97.7} & 58.6 & 86.4 & \textbf{86.8} & \textbf{94.3} & \textbf{85.0}  \\
        
         \bottomrule
         \end{tabular}
     }
     \caption{
         Safety performance comparison, highlighting \modelname{}.
     }
     \label{tab:safety_evaluation}
 \end{table*}

\section*{Conclusion}
In this work, we present \modelname{}, a family of code LLMs that advance the state-of-the-art in autonomous software engineering through the code-flow pre-training paradigm and multi-phase evolutionary training. By capturing dynamic repository transitions and integrating extensive reasoning trajectories with repository-scale context during mid-training, our models establish robust logical foundations for complex code intelligence tasks. \modelname{} demonstrates exceptional performance across diverse benchmarks spanning agentic software engineering, competitive programming, and tool use, validating the effectiveness of our training methodology. The \modelname{} (Loop variant) further addresses practical deployment challenges through recurrent architectural innovations that optimize the capacity-efficiency trade-off. We deliver specialized models tailored for both deep analytical reasoning and general assistance scenarios. By open-sourcing the complete training pipeline and model checkpoints, we aim to catalyze further research in code intelligence and accelerate the development of production-ready agentic systems capable of tackling real-world software engineering challenges.

\clearpage
\newpage

\section{Contributions and Acknowledgements}
\textbf{The authors of this paper are listed in order as follows:} \\
Jian Yang, Wei Zhang, Shawn Guo, Zhengmao Ye, Lin Jing, Shark Liu, Yizhi Li, Jiajun Wu, Cening Liu, X. Ma, Yuyang Song, Siwei Wu, Yuwen Li, L. Liao, T. Zheng, Ziling Huang, Zelong Huang, Che Liu, Yan Xing, Renyuan Li, Qingsong Cai, Hanxu Yan, Siyue Wang, Shikai Li, Jason Klein Liu, An Huang, Yongsheng Kang, Jinxing Zhang, Chuan Hao, Haowen Wang, Weicheng Gu, Ran Tao, Mingjie Tang, Peihao Wu, Jianzhou Wang, Xianglong Liu, Weifeng Lv, Bryan Dai.

\paragraph{Core Contributors:} Jian Yang, Wei Zhang, Shawn Guo, Zhengmao Ye, Lin Jing, Shark Liu, Yizhi Li, Jiajun Wu.

\paragraph{Contributors} Cening Liu, Xi Lin, Yuyang Song, Siwei Wu, Yuwen Li, L. Liao, Tianyu Zheng, Ziling Huang, Zelong Huang, Che Liu, Yan Xing, Renyuan Li, Qingsong Cai, Hanxu Yan, Siyue Wang, Shikai Li, Jason Klein Liu, An Huang, Yongsheng Kang, Jinxing Zhang, Chuan Hao, Jing Yang, Haowen Wang, Weicheng Gu, IQuest Coder.

\paragraph{Leadership and Senior Advisory Committee:} Ran Tao, Mingjie Tang, Peihao Wu, Jianzhou Wang, Xianglong Liu, Weifeng Lv.

\paragraph{Corresponding Authors:} Bryan Dai.

\bibliography{ref}

\end{document}